\documentclass[runningheads]{llncs}
\usepackage{marvosym}
\usepackage{authblk}
\usepackage{graphicx}
\usepackage{epstopdf}
\usepackage{booktabs}
\usepackage{diagbox}  
\usepackage{multirow}
\usepackage{amstext}
\usepackage{hyperref}
\usepackage{amsfonts, mathtools}
\newdimen\Lmargin
\newdimen\Rmargin
\Rmargin=\paperwidth \advance\Rmargin by-\Lmargin \advance\Rmargin by-\textwidth
\ifdim\Lmargin>\Rmargin \Rmargin=\Lmargin \fi

\begin{document}

\author{
Zhengyang Li  \inst{1}
\and
Wenhao Liang  \inst{1}
\and 
Chang Dong\thanks{Email: chang.dong@adelaide.edu.au} \inst{1}
\and
Weitong Chen  \inst{1}
\and
Dong Huang  \inst{2}
}

\institute{
School of Computer and Mathematical Sciences, University of Adelaide, 5000, SA, Australia 
\and
Institute of Plasma Physics and Technology, Jiangsu Key Laboratory of Thin Films, School of Physical Science and Technology, Soochow University, Suzhou 215006, China
}

\title{Correlation Analysis of Adversarial Attack in Time Series Classification}

%
%

\maketitle 
\begin{abstract}

    This study investigates the vulnerability of time series classification models to adversarial attacks, with a focus on how these models process local versus global information under such conditions. By leveraging the Normalized Auto Correlation Function (NACF), an exploration into the inclination of neural networks is conducted. It is demonstrated that regularization techniques, particularly those employing Fast Fourier Transform (FFT) methods and targeting frequency components of perturbations, markedly enhance the effectiveness of attacks. Meanwhile, the defense strategies, like noise introduction and Gaussian filtering, are shown to significantly lower the Attack Success Rate (ASR), with approaches based on noise introducing notably effective in countering high-frequency distortions. Furthermore, models designed to prioritize global information are revealed to possess greater resistance to adversarial manipulations. These results underline the importance of designing attack and defense mechanisms, informed by frequency domain analysis, as a means to considerably reinforce the resilience of neural network models against adversarial threats.

    \keywords{Time Series Classification \and Correlation Function \and Local vs Global Information preference \and Adversarial Attack vs Defense.}
\end{abstract}

\vspace{-0.8cm}
\section{Introduction}
\vspace{-0.3cm}

Recently, time series data has become increasingly prevalent. As we transit into the era of Industry 4.0, countless sensors generate vast volumes of time series data~\cite{Fawaz2019,karim2021adversarial,siddiqui2020benchmarking}. Correspondingly, the application of deep neural networks (DNNs) for time series classification (TSC) has surged in popularity~\cite{Fawaz2019,Ismail_Fawaz_2020}. However, DNNs exhibit vulnerabilities to minor perturbations in input data, often leading to misclassification, indicating low resistance to external perturbation, which has attracted significant attention from the research community across various domains~\cite{brendel2018decisionbased,szegedy2014intriguing,madry2019deep,goodfellow2015explaining,Wang2021SSADR}.

\begin{figure}
    \centering
    \includegraphics[width=0.9\textwidth]{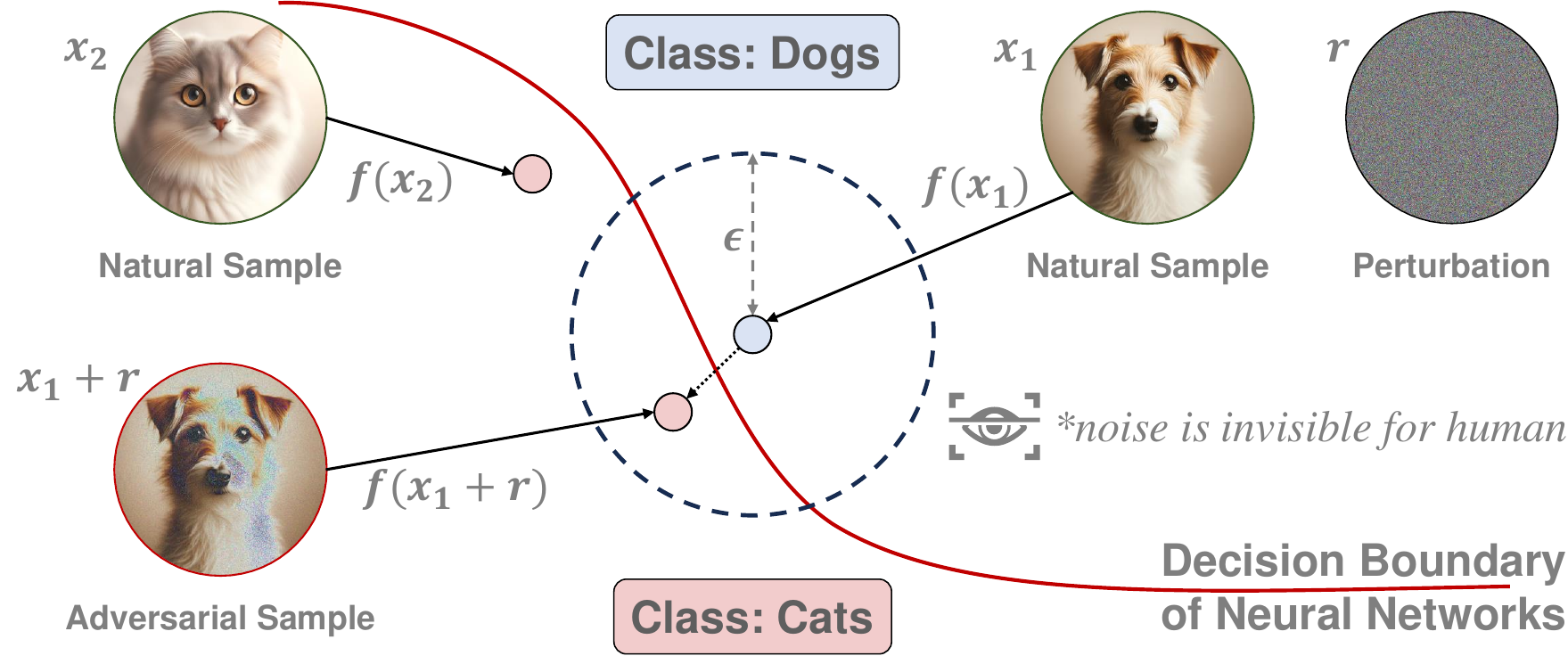}
    \vspace{-0.2cm}
    \caption{A diagram illustrating the susceptibility of neural networks to external noise. In this diagram, $f$ represents the model, $x_1$ and $x_2$ are samples, $r$ denotes the perturbation, and $\epsilon$ signifies the maximum allowable magnitude of $r$.} \label{fig1}
\end{figure}

As illustrated in Figure~\ref{fig1}, even a minor change to the input image can cause a Deep Neural Network (DNN) to misclassify across its decision boundary, which underscores the vulnerability of neural networks to external noise. Such errors raise considerable concerns and risks, especially when implementing these advanced technologies in real-world scenarios~\cite{wang2021adversarial}, which has garnered substantial evidence within the field of computer vision~(CV). For example, a ``Stop Sign" can be attacked as ``Speed Limit" by adding specific patterns on the surface of the sign, while it can still be recognized as a stop sign in human eyes ~\cite{Olszewski2001Generalized}, which can lead to a severe consequence in the self-driving automobile.

Moreover, time series data, which have been endorsed as an important research aspect in medical, financial and engineering domains~\cite{lim2021time}, also suffer from signal distortions during transmission or deliberate manipulations by malicious entities or noise generated by the environment. Given that many TSC models are pivotal in medical sectors~\cite{han2020deep,shao2022defending,www2023advattack}, the consequences of successful adversarial attacks can be even more devastating. Therefore, researchers are striving to devise robust defense against these malevolent activities by developing techniques of adversarial attacks to detect the faults of TSC models.

Most algorithms for TSC attacks have been directly adapted from those in CV. For instance, Fast Gradient Sign Method (FGSM)~\cite{goodfellow2015explaining}  can achieve successful but easily detectable attacks due to the distinctive saw-tooth pattern in the time series. To mitigate this, Gautier. et. al. introduced regularization as a Smooth Gradient Method(SGM) to enhance the stealthiness of the attacks~\cite{pialla2022smooth}, albeit at the expense of the Attack Success Rate (ASR). Chang et al.~\cite{Dong2023} proposed a SWAP attack, aiming to improve the Attack Success Rate (ASR) while minimizing the distance level, by adjusting logits. Despite their effectiveness, these strategies-originating from CV—do not fully illuminate the impact of adversarial patterns on model predictions within TSC.

To bridge this gap, our study turns to the traditional signal processing tool: the correlation function, renowned for its capability to examine a system's response to external stimuli. Here, the correlation function is applied as a regularization term in adversarial attacks to thoroughly investigate how neural networks differentiate between local and global information during training and their consequent vulnerability to adversarial threats. Specifically, this paper employs the Normalized Auto Correlation Function (NACF) to dissect neural networks' tendencies, marking a pivotal contribution to understanding and enhancing TSC model resilience against adversarial attacks.The main contribution of this work can be summarised as:

\begin{itemize}
    \item[$\bullet$]
        It benchmarked an analysis of adversarial perturbations' impact on time series classification models and assesses the defense strategies, thereby broadening the understanding of model vulnerabilities and mitigation techniques.

    \item[$\bullet$]
        Utilizing the Normalized Auto Correlation Function (NACF), the research reveals neural networks' propensity for prioritizing local over global information under adversarial conditions, contributing novel insights into model biases and processing tendencies.

    \item[$\bullet$]
        The effectiveness of frequency-focused regularization techniques, particularly those employing Fast Fourier Transform (FFT) methods, and the superior resilience of models emphasizing global information processing, underscore the importance of frequency domain analysis in developing robust defense mechanisms.
\end{itemize}

\vspace{-0.4cm}
\section{Related Works}\label{sec:related_works}

\vspace{-0.2cm}
\subsection{Preliminary of Adversarial Attack}

In time series classification, an adversarial attack refers to a malicious attempt to introduce slight perturbations to a time series \( x \in \mathbb{R}^d \) to produce a closely related series $x'=x+r$~(\( x' \in \mathbb{R}^d \)) with the goal of altering the predicted label. This can be mathematically characterized by the equation:

\begin{equation}
    \text{argmax}\ \{f(x)\} \ne \text{argmax}\ \{f(x + r)\},  \text{s.t.} \ ||r||^2 \ll ||x||^2.
\end{equation}
Here, \( f(x) \) represents the predicted probability distribution over the labels for the input \( x \)
and perturbation \( r \) is intentionally small in magnitude relative to \( x \) as indicated by their norms. In GM based methods, including SWAP, the perturbed logits and the target logits are used as the main part of loss function to measure the difference between the two distributions.

\vspace{-0.2cm}
\subsection{Adversarial Attacks}
There are main stream of adversarial attacks: White-Box and Black-Box. White-Box attacks, where the attacker has access to the model's architecture, dataset, and parameters, facilitate gradient-based strategies. In contrast, Black-Box attacks operate with limited information, making them more challenging. Most of attacking methods are developed as the white-box attacks targeting image classification task to examine the trustworthiness of model. Goodfellow et al.~\cite{goodfellow2015explaining} introduced the Fast Gradient Sign Method (FGSM), a straightforward approach that generates perturbations by moving in the gradient direction with a small scalar. Madry et al.~\cite{madry2019deep} expanded on this with Projected Gradient Descent (PGD), an iterative version of FGSM. Similarly, Kurakrin et al.~\cite{pialla2022smooth} proposed the Basic Iterative Method (BIM), to seek perturbations that maximize the loss for a input data while  the noise amplitude is clipped under a threshold ($\epsilon$) to conceal the obvious perturbations. Carlini et al.~\cite{carlini2017towards} framed the attack as an optimization problem aiming to minimize the distance (e.g. $\mathcal{L}_2$ norm) between original sample and adversarial sample upon a successful attack (C\&W method). Notably, SWAP highlights the use of KL-divergence and the strategic selection of labels to challenge the original prediction effectively~\cite{Dong2023}. This targeted label approach, as evidenced in preliminary experiments, yields higher ASR compared to random target selection methods like PGD.

\vspace{-0.2cm}
\subsection{Adversarial Defense}

On the defense side, Papernot et al.~\cite{papernot2016distillation} introduced a distillation-based approach to transfer knowledge from a complex network to simpler ones, enhancing resilience to adversarial attacks and improving generalization against adversarial samples. Madry et al.~\cite{madry2019deep} developed an adversarial training method, framing it as a min-max optimization problem that integrates both adversarial and original samples during training to enhance model robustness. Expanding on this, Kannan et al.~\cite{kannan2018logit} employed a logit pairing technique to minimize the difference in logits between original and adversarial samples, adding a regularizer to the training process. Similarly, Ma et al.~\cite{ma2020soar} approached defense by minimizing empirical risk through adversarial training, introducing a second-order adversarial regularizer based on a Taylor series expansion for robustness improvement. Additionally, Sammangouei et al.~\cite{samangouei2018defense} utilized a Wasserstein GAN framework for adversarial training, aiming to reconstruct original images from adversarial samples by minimizing reconstruction errors.

\vspace{-0.4cm}
\section{Proposed Method}\label{sec:proposed_method}

\vspace{-0.2cm}
\subsection{Theoretical Analysis}
\vspace{-0.2cm}
\subsubsection{Similarity metriced by the Correlation Function.}

Suppose we have two infinitely long discrete time series, $x$ and $x'=x+r$, where $r$ represents a residual or noise sequence. We aim to calculate their degree of similarity.
\begin{equation}
    \label{correlation}
    \text{Time-Lagged Correlation}_{xx'}(\tau) = \langle(x(t) - \bar{x}) (x'(t+\tau) - \bar{x'})\rangle.
\end{equation}
In this equation, $\tau$ represents the time delay, a parameter that allows us to explore the correlation between the sequences at various intervals. Specifically, $x'(t)$ denotes the value of the first time series at time point $t$, and $x'(t+\tau)$ corresponds to the value of the second time series at time point $t+\tau$. This latter value can be seen as the first series displaced by a time delay $\tau$, potentially incorporating noise or other modifications. The symbols $\bar{x}$ and $\bar{x'}$ represent the mean values of the time series $x(t)$ and $x'(t)$, respectively. This formulation captures the essence of how the similarity between two time series evolves as they are shifted relative to each other over time. By summing over the time series, we can obtain a measure of the similarity between the two time series:
\begin{equation}
    \label{correlation_sum}
    \text{Correlation Sum}_{xx'} = \lim_{N\rightarrow\infty} \frac{1}{N} \sum_{\tau=0}^{N-1} (x(t) - \bar{x}) (x'(t+\tau) - \bar{x'}).
\end{equation}
Here $N$ is the length of the time series. Then we introduce a weight function $w(\tau,k)$ during the summation process:
\begin{equation}
    \label{weight}
    w(\tau,k) = \frac{(1 + e^{-(\tau - k)})^{-1}}{\sum_{\tau=0}^{N-1} (1 + e^{-(\tau - k)})^{-1}},
\end{equation}
where the parameter $k$, referred as the "midpoint", is the critical value at which the Sigmoid function reaches a value of $1/2$. By incorporating the weight function $w(t)$ into the similarity calculation during the summation process, we derive a weighted similarity measure:
\begin{equation}
    \label{weighted_correlation}
    \text{Weighted Corr Sim}_{xx'}(k) = \lim_{N\rightarrow\infty} \sum_{\tau=0}^{N-1} w(\tau,k) \cdot (x(t) - \bar{x}) (x'(t+\tau) - \bar{x'}).
\end{equation}

This Weighted Correlation Similarity measure can be incorporated into the overarching loss function of a neural network, serving as a regularization component alongside the cross-entropy:
\begin{equation}
    \mathcal{L} = H(P, Q) + a \cdot \text{Weighted Corr Sim}_{xx'}(k),
    \label{wcs_cross_entropy}
\end{equation}
where $H(P, Q)$ represents the cross-entropy loss and $a$ is weight coefficient to balance. The adaptation of the midpoint parameter $k$ provides an additional lever to fine-tune the attention of neural network, in adversarial attack, towards either more local or global features within the time series, depending on the specific requirements of the application.

This mathematical framework, inspired by principles from the linear response theory~\cite{Huang2022OViscosity}, draws a parallel between the dynamic behavior of physical systems under perturbation and the response of neural networks to adversarial inputs. In essence, it provides a robust theoretical foundation for leveraging time-lagged correlations as a metric for assessing and enhancing the robustness of machine learning models against adversarial attacks, grounding the approach in well-established scientific concepts.

\vspace{-0.2cm}
\subsubsection{\textbf{Pre-experiment.}}

In this preliminary experiment, by using weighted similarity~\ref{weighted_correlation} as regularization, we investigate the influence of the sigmoid function's midpoint adjustment on the model's vulnerability to adversarial attacks. By conducting an empirical study across 128 datasets using the Inception Time model, we explored the relationship between the midpoint position of the sigmoid function and the attack success rate (ASR).

\begin{figure}[h]
    \centering
    \includegraphics[width=0.7\textwidth]{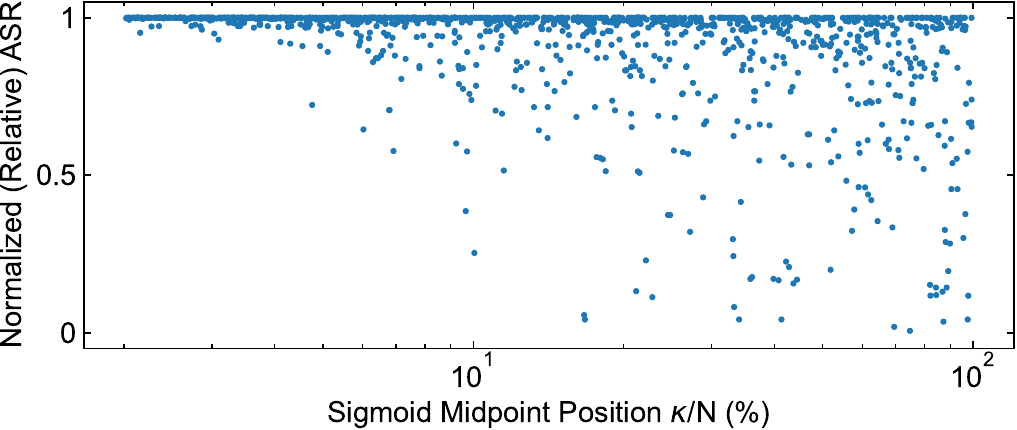}
    \vspace{-0.2cm}
    \caption{This scatter plot visualizes the relationship between the sigmoid function's midpoint position, expressed as a percentage of its range (X-axis), and the Normalized Attack Success Rate (Relative ASR) (Y-axis). Here, the Y-axis values are normalized to the highest ASR observed across trials for the same dataset, but with varying midpoints~$k$ of the sigmoid function.} \label{fig3}
\end{figure}

For each dataset, after training the model, we randomly selected ten midpoints in logarithmic space and subjected the trained models to attacks, calculating the ASR at each midpoint. Given the significant variance in ASR across datasets, we normalized the ASRs within each dataset against the highest ASR observed, obtaining a relative ASR. This normalization allowed us to control the neural network's focus on specific scales of the time series by adjusting the midpoint. Statistically, in fig~\ref{fig3}, we observed a trend where an increase in the midpoint led to a gradual decline in the relative ASR, suggesting that the model, during training, tends to focus more on local (small $\tau$, high-frequency) information rather than global (large $\tau$, low-frequency) information.

Based on these findings, we hypothesize that introducing a regularization mechanism that increases the frequency component of $r$ could enhance the effectiveness of attacks. Conversely, encouraging the model to focus more on global information during training could improve its robustness. To test this hypothesis, we devised two regularization methods aimed at increasing the frequency of $r$:

\begin{itemize}
    \item[$\bullet$]
        Applying a fast Fourier transform (FFT) to $r$, multiplying it by a function similar to the sigmoid used previously, and then performing an inverse FFT. This transformed $r$ with its associated weight is then used as a regularization term in the loss function. Let the Fast Fourier Transform and its inverse be denoted by $\text{fft}$ and $\text{ifft}$, respectively. By using the low frequency-based regularization term, the following loss can enfore the perturbation $r$ has more components on high frequency:

        \begin{equation}
            \mathcal{L}_\text{FFT} = H(P, Q) + a_1 \cdot \text{FFT}^{-1}(\text{FFT}(r) \cdot w(t)).
            \label{FFT_loss}
        \end{equation}

    \item[$\bullet$]
        Starting from the concept of correlation and focusing on the limit case where $\tau=0$, the correlation of $x$ and $y$ ($y=x+r$) simplifies to the cosine similarity between $x$ and $y$. By maximizing this cosine similarity, we aim to increase the high-frequency components of $r$.

        \begin{equation}
            \mathcal{L}_\text{COS} = H(P, Q) + a_2 \cdot \log_{10}\left(\frac{x \cdot x'}{|x||x'|} + 1\right).
            \label{FFT_COS}
        \end{equation}
\end{itemize}
Here, $H(P, Q)$ represents the cross-entropy loss, $a_1$ and $a_2$~($a_2<0$) are weight coefficients to balance the regularization term and the cross-entropy loss.

These approaches are grounded in the observation that models trained with a focus on capturing global information exhibit increased robustness to adversarial attacks, suggesting a promising direction for enhancing model security against such threats.

\vspace{-0.2cm}
\subsection{Attack and Defense Framework}
\vspace{-0.2cm}

\begin{figure}
    \centering
    \includegraphics[width=1\textwidth]{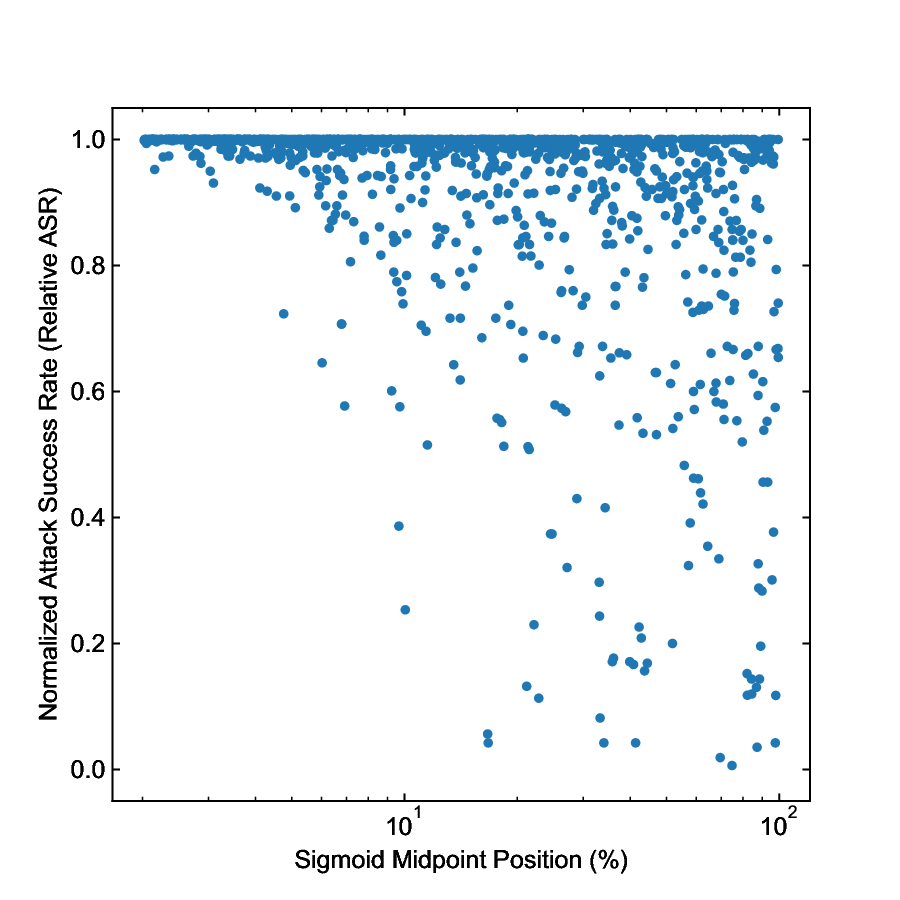}
    \vspace{-0.2cm}
    \caption{proposed Attack and Defense
        Framework For Time Series Classification models. Note that, the green pathway shows how traditional model trained, and the red pathway explains how the perturbation $r$ learned by the framework.} \label{fig2}
\end{figure}

Figure~\ref{fig2} elucidates the structure of our comprehensive attack and defense framework. During the forward propagation process, an input data first traverses a defense layer, where it encounters three paths: direct passage without any augmentation; application of Gaussian filtering for smoothing; or the introduction of white noise before being forwarded to the neural network. For conventional classification tasks, a neural network typically consists of an encoder, a fully connected layer, and a softmax layer. In this paper, we explore five distinct models, each representing a different mainstream neural network architecture, with specifics to be detailed later. Following the passage through the fully connected layer, the cross-entropy between the predicted and actual labels is computed and backpropagated. In adversarial attacks, the objective is to minimize the cross-entropy between the predicted distribution and a desired distribution \( P \), a concept derived from the\cite{Dong2023} paper, which involves swapping labels of the highest and second-highest probabilities. During back propagation, rather than updating the model \( f \), the perturbation \( r \) is adjusted, and the modified input \( x+r \) is fed back into the network. This architectural design allows us to examine the efficacy and drawbacks of various attack and defense strategies, as well as assess different models. The results of these experiments are detailed in the following section.

\vspace{-0.4cm}
\section{Experiments and Results}\label{sec:experiments_and_results}

\vspace{-0.2cm}
\subsection{Experimental Setup}

\vspace{-0.2cm}
\subsubsection{Dataset.}

To validate the effectiveness of attack and defense to the TSC Model, the UCR2018 datasets~\cite{UCRArchive2018} were applied to train the model. These 128 datasets span a diverse range of real-world domains, including healthcare, agriculture, finance, engineering and more. Each dataset comprises a distinct number of samples, all of which have been pre-partitioned into training and testing sets.

\vspace{-0.2cm}
\subsubsection{\textbf{Models.}}

Here, we use five different models, namely Inception Time v4~\cite{Dong2023}, LSTM-FCN, MACNN, ResNet18, TS2V~\cite{yue2022ts2vec}. These five models represent several different neural network architectures that are currently mainstream.

\vspace{-0.2cm}
\subsubsection{\textbf{Evaluation Metrics.}}

For a rigorous and equitable comparison with existing methodologies, the Attack Success Rate (ASR) and the Mean Success Distance (MSD, based on the $\mathcal{L}^2$ Distance) are used as our metrics for evaluating the performance of attack and defense mechanisms. ASR quantifies the effectiveness of an attack by calculating the ratio of misclassified adversarial samples to the total adversarial samples generated. Meanwhile, MSD measures the effectiveness and subtlety of an attack by averaging the $\mathcal{L}^2$ distance between original and corresponding adversarial samples.

\vspace{-0.2cm}
\subsubsection{\textbf{Implementation Details.}}

Our model was developed using PyTorch 2.0, diverging from the Tensorflow-based baseline~\cite{Dong2023}, where the training parameters remained consistent. Our experiments were executed on a server outfitted with Nvidia RTX 4090 GPUs, 128 GB RAM, and Dual Xeon E5-2667 v4 processors. We standardized training across all models to 1000 epochs, while adversarial attacks were assessed 100 epochs to gauge their impact.

\vspace{-0.2cm}
\subsection{Attack Performance Comparison}
\vspace{-0.1cm}
\begin{table}[h]
    \centering
    \caption{Attack Performance Comparison on Models Without Defense}
    \vspace{-0.2cm}
    \label{table1}
    {\small
        \begin{tabular}{lllllllllll}
            \toprule
            \multirow{2}{*}{\diagbox[dir=NW]{\fontsize{6pt}{1pt}\selectfont Model}{\fontsize{6pt}{1pt}\selectfont  Attack}} & \multicolumn{2}{c}{PGD} & \multicolumn{2}{c}{SWAP} & \multicolumn{2}{c}{SWAP($\mathcal{L}^2$)} & \multicolumn{2}{c}{COS} & \multicolumn{2}{c}{FFT}                                                                    \\
            \cmidrule(r){2-3} \cmidrule(r){4-5} \cmidrule(r){6-7} \cmidrule(r){8-9} \cmidrule(r){10-11}
                                                                                                                            & ASR                     & MSD                      & ASR                                       & MSD                     & ASR                     & MSD            & ASR   & MSD   & ASR            & MSD            \\
            \midrule
            Inception                                                                                                       & 0.648                   & 1.325                    & 0.736                                     & 1.060                   & 0.735                   & 0.935          & 0.736 & 1.064 & \textbf{0.738} & \textbf{0.740} \\
            LSTMFCN                                                                                                         & 0.639                   & 1.373                    & 0.726                                     & 1.150                   & 0.723                   & 0.979          & 0.726 & 1.157 & \textbf{0.729} & \textbf{0.645} \\
            MACNN                                                                                                           & 0.513                   & 0.697                    & 0.580                                     & 0.831                   & 0.402                   & \textbf{0.262} & 0.581 & 0.836 & \textbf{0.594} & 0.489          \\
            ResNet18                                                                                                        & 0.869                   & 0.656                    & 0.907                                     & 0.501                   & 0.906                   & 0.431          & 0.907 & 0.505 & \textbf{0.908} & \textbf{0.323} \\
            TS2V                                                                                                            & 0.508                   & 1.275                    & 0.651                                     & 1.361                   & 0.441                   & \textbf{0.541} & 0.652 & 1.365 & \textbf{0.653} & 0.855          \\
            \bottomrule
        \end{tabular}
    }
\end{table}
\vspace{-0.2cm}

We begin our discussion with Table~\ref{table1}, where we evaluate five adversarial attack methods: PGD, SWAP, SWAP($\mathcal{L}^2$), COS, and FFT, across five neural network architectures to assess the universality of our results. Notably, the FFT approach, which increases the frequency component \(r\), consistently achieves higher ASR and lower MSD in the majority of cases. Conversely, the SWAP($\mathcal{L}^2$) regularization method exhibits a significant decline in ASR when MSD values are low, suggesting potential limitations in its effectiveness. Meanwhile, the efficacy of COS regularization remains less evident, possibly due to suboptimal parameter settings, highlighting an avenue for further experimentation.

Our custom implementation of SWAP notably outperforms the Classic Gradient Method~(PGD), affirming the validity of the SWAP approach and reinforcing the reliability of our experimental outcomes. This study deliberately omits comparisons with other methods, focusing instead on validating our hypothesis: that perturbed \(x\) focus on local correlation~(higher-frequency) enhances attack success. This is further substantiated by introducing regularization to SWAP~(as opposed to PGD), allowing for controlled variable analysis. Our observations reveal that PGD can lead to unstable ASR values across multi-label datasets, attributing to its decision-space directionality. In contrast, SWAP demonstrates remarkable stability, underscoring the strategic advantage of our chosen methodology.

\vspace{-0.4cm}
\begin{figure*}[h]
    \centering
    \includegraphics[width=1\textwidth]{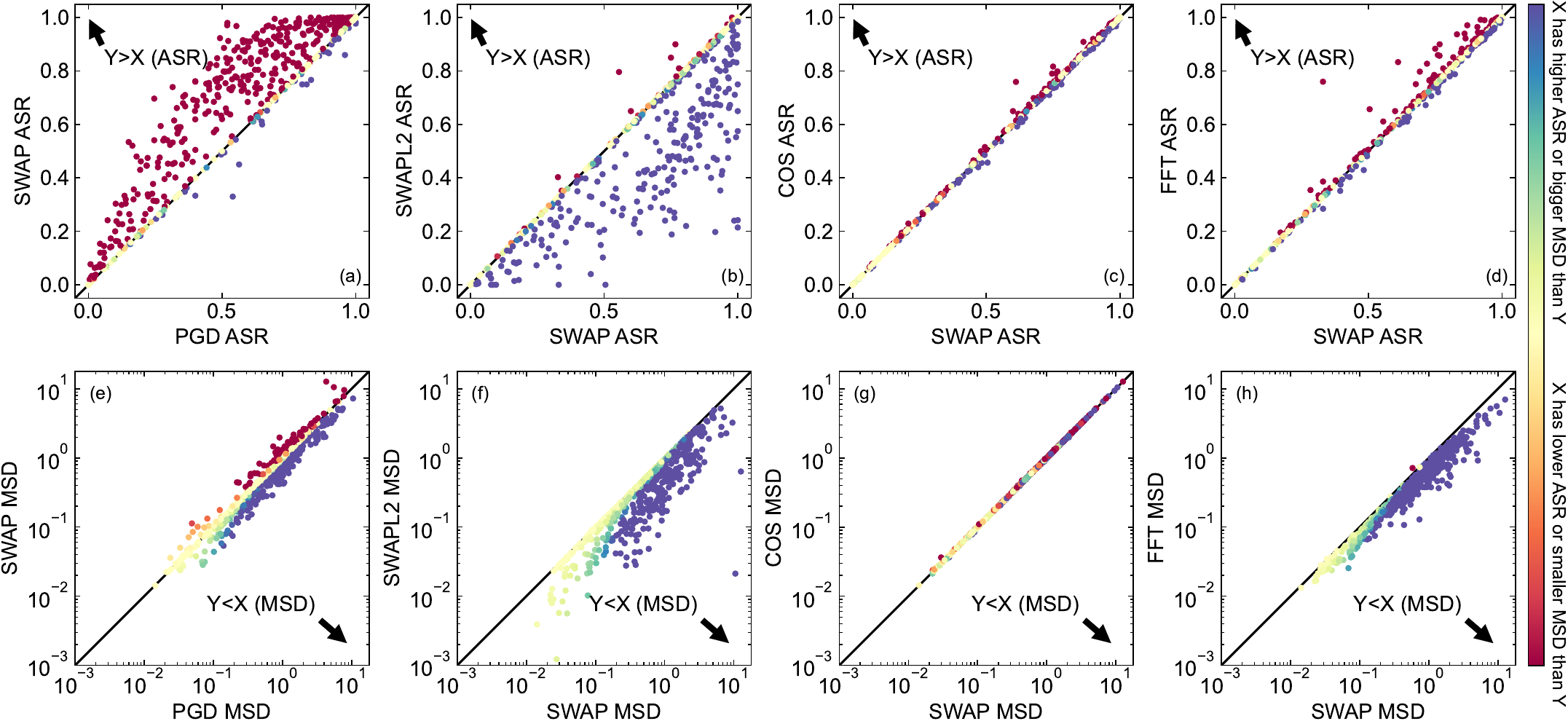}
    \vspace{-0.2cm}
    \caption{Comparison of Attack Success Rate (ASR) and Mean Success Distance (MSD) among PGD (GM), SWAP, SWAP($l^2$), COS, and FFT algorithms.}\label{fig4}
\end{figure*}

In Figure~\ref{fig4}, we evaluate the efficacy and subtleties of five adversarial attack methodologies. This figure meticulously contrasts the Attack Success Rate (ASR) and Mean Success Distance (MSD) across these methods. A data point positioned above the $y=x$ line in the ASR comparison signifies superior performance of the corresponding Y-axis method over its X-axis counterpart. Conversely, for the MSD comparison, a point below the $y=x$ line indicates a more effective minimization of perturbation by the Y-axis method. Specifically, the comparisons between SWAP and PGD (GM) across subfigures (a) and (e) indicate SWAP has its superiority in ASR across numerous datasets without a significant reduction in MSD, hinting at the influence of target label chosen on the results. Moreover, the slight advantage of COS over SWAP in ASR, as shown in subfigures (c) and (g), alongside almost identical MSD values, underscores the potential unrealized potential of COS regularization, necessitating further experimental validations. The FFT method, as evaluated in subfigures (d) and (h), not only surpasses SWAP in ASR across the majority of datasets but also demonstrates a lower MSD, aligning with our initial hypothesis regarding the effectiveness of higher-frequency perturbations. Our observation reveals that FFT, by emphasizing higher frequency modifications in \(r\), consistently achieves better ASR and lower MSD across most datasets, suggesting its efficiency in generating subtle and successful adversarial samples.

\vspace{-0.2cm}
\subsection{Defense Performance Comparison}
\vspace{-0.4cm}

\begin{figure*}[h]
    \centering
    \includegraphics[width=0.97\textwidth]{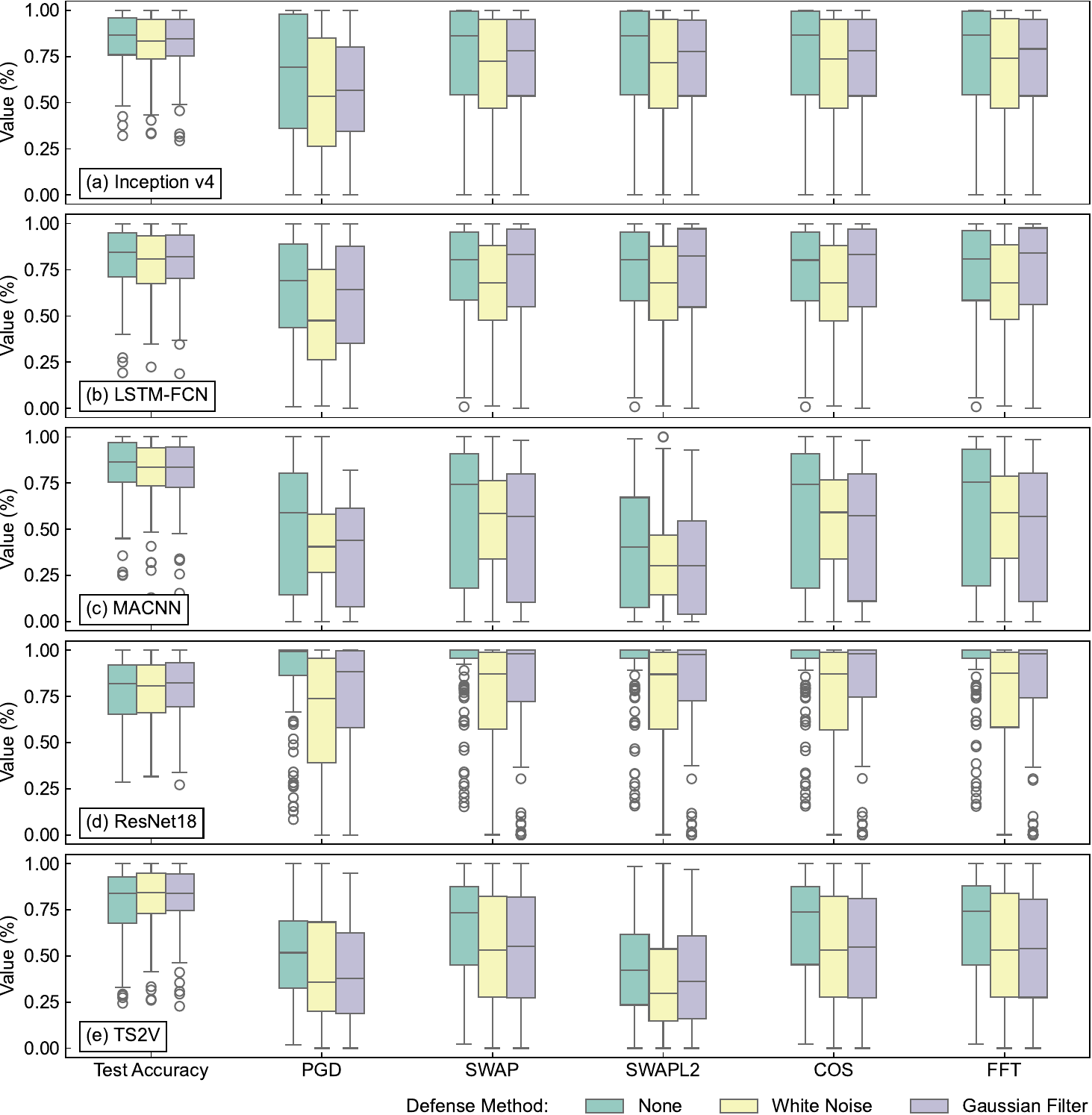}
    \vspace{-0.2cm}
    \caption{This box plot presents comprehensive experimental results across selected models, each embodying a mainstream neural network architecture. Distinct colors denote various defense methods. For each subplot, the initial column illustrates the test accuracy, while the subsequent five columns detail the Attack Success Rate (ASR) against five distinct attack methodologies.
    } \label{fig5}
\end{figure*}

In our subsequent analysis, we assess the efficacy of employing noise and Gaussian filtering as defense mechanisms. Figure~\ref{fig5} encapsulates the entirety of our experimental data, spanning from subgraphs (a) to (e), each delineating different models. Within each subplot, the leftmost column delineates the model's test accuracy, followed by five columns representing the ASR for diverse attack methodologies. Distinct coloration signifies varied attack methods. A preliminary examination of test accuracy reveals that lighter models such as Inception v4, LSTM-FCN, and MACNN exhibit a decrement in test performance upon the integration of noise as a defensive strategy. Intriguingly, the TS2V model demonstrates a significant augmentation in test accuracy when training incorporates augmentation, a phenomenon warranting further discussion. Moreover, our observations indicate that models fortified with Gaussian smoothing (filtering) slightly outperform their counterparts employing noise in terms of test set performance. This suggests that discarding high-frequency information exerts a lesser adverse impact on data integrity compared to the substitution of erroneous high-frequency information.

\vspace{-0.4cm}
\begin{table}[h]
    \centering
    \caption{Comparison on Models With White Noise Augmentation}
    \vspace{-0.2cm}
    \label{table2}
    {\small
        \begin{tabular}{lllllllllll}
            \toprule
            \multirow{2}{*}{\diagbox[dir=NW]{\fontsize{6pt}{1pt}\selectfont Model}{\fontsize{6pt}{1pt}\selectfont  Attack}} & \multicolumn{2}{c}{PGD} & \multicolumn{2}{c}{SWAP} & \multicolumn{2}{c}{SWAP($\mathcal{L}^2$)} & \multicolumn{2}{c}{COS} & \multicolumn{2}{c}{FFT}                                                                             \\
            \cmidrule(r){2-3} \cmidrule(r){4-5} \cmidrule(r){6-7} \cmidrule(r){8-9} \cmidrule(r){10-11}
                                                                                                                            & ASR                     & MSD                      & ASR                                       & MSD                     & ASR                     & MSD            & ASR            & MSD   & ASR            & MSD            \\
            \midrule
            Inception                                                                                                       & 0.540                   & 1.984                    & 0.665                                     & 1.550                   & 0.662                   & 1.345          & 0.664          & 1.554 & \textbf{0.667} & \textbf{1.019} \\
            LSTMFCN                                                                                                         & 0.512                   & 2.442                    & 0.644                                     & 2.064                   & 0.641                   & 1.716          & 0.644          & 2.070 & \textbf{0.645} & \textbf{1.084} \\
            MACNN                                                                                                           & 0.430                   & 1.176                    & 0.559                                     & 1.453                   & 0.339                   & \textbf{0.425} & 0.560          & 1.465 & \textbf{0.564} & 0.967          \\
            ResNet18                                                                                                        & 0.659                   & 1.749                    & 0.762                                     & 1.296                   & 0.761                   & 1.127          & 0.761          & 1.297 & \textbf{0.763} & \textbf{0.922} \\
            TS2V                                                                                                            & 0.420                   & 2.138                    & 0.540                                     & 2.158                   & 0.359                   & \textbf{0.779} & \textbf{0.542} & 2.168 & \textbf{0.542} & 1.307          \\
            \bottomrule
        \end{tabular}
    }
\end{table}

\vspace{-1cm}
\begin{table}[h]
    \centering
    \caption{Comparison on Models With Gaussian Smooth Augmentation}
    \vspace{-0.2cm}
    \label{table3}
    {\small
        \begin{tabular}{lllllllllll}
            \toprule
            \multirow{2}{*}{\diagbox[dir=NW]{\fontsize{6pt}{1pt}\selectfont Model}{\fontsize{6pt}{1pt}\selectfont  Attack}} & \multicolumn{2}{c}{PGD} & \multicolumn{2}{c}{SWAP} & \multicolumn{2}{c}{SWAP($\mathcal{L}^2$)} & \multicolumn{2}{c}{COS} & \multicolumn{2}{c}{FFT}                                                                    \\
            \cmidrule(r){2-3} \cmidrule(r){4-5} \cmidrule(r){6-7} \cmidrule(r){8-9} \cmidrule(r){10-11}
                                                                                                                            & ASR                     & MSD                      & ASR                                       & MSD                     & ASR                     & MSD            & ASR   & MSD   & ASR            & MSD            \\
            \midrule
            Inception                                                                                                       & 0.568                   & 1.282                    & 0.693                                     & 0.936                   & 0.691                   & 0.817          & 0.692 & 0.938 & \textbf{0.698} & \textbf{0.615} \\
            LSTMFCN                                                                                                         & 0.599                   & 1.059                    & 0.713                                     & 0.851                   & 0.711                   & 0.732          & 0.713 & 0.856 & \textbf{0.717} & \textbf{0.514} \\
            MACNN                                                                                                           & 0.373                   & 0.775                    & 0.481                                     & 0.930                   & 0.321                   & \textbf{0.283} & 0.484 & 0.947 & \textbf{0.486} & 0.612          \\
            ResNet18                                                                                                        & 0.737                   & 0.969                    & 0.807                                     & 0.715                   & 0.807                   & 0.654          & 0.807 & 0.721 & \textbf{0.810} & \textbf{0.601} \\
            TS2V                                                                                                            & 0.400                   & 1.878                    & \textbf{0.525}                            & 1.883                   & 0.387                   & \textbf{0.783} & 0.522 & 1.880 & 0.524          & 1.117          \\
            \bottomrule
        \end{tabular}
    }
\end{table}
\vspace{-0.3cm}

Shifting focus to the impact of defense implementations on attack outcomes, the rightmost five columns correspond to the previously discussed five attack methodologies. For contextual comparison, ASR values without defense implementation are also depicted. Observations indicate a general reduction in ASR across most models upon defense integration, affirming the efficacy of the defense strategies deployed. Notably, models fortified with noise as a defense mechanism exhibited a more pronounced reduction in ASR, suggesting a greater impact of introducing incorrect high-frequency information during model training than the absence of specific model weight adjustments (pertaining to the decision space distribution). To furnish quantifiable insights, Tables~\ref{table2} and \ref{table3} delineate the ASR and MSD values post-application of noise and Gaussian smoothing as defensive measures. A comparative analysis with Table~\ref{table1} reveals a decline in ASR and an elevation in MSD upon incorporating defense layers, aligning with the findings illustrated in Figure~\ref{fig5}.

In the TS2V model specifically, the introduction of Gaussian filtering led to a decrease in the ASR of the FFT attack method, falling below its default performance. This phenomenon is attributed to the Gaussian filter's propensity to eliminate high-frequency information, which is precisely the domain FFT attack methods exploit to craft adversarial samples. Remarkably, TS2V's resilience can be seen as evidence that learning global information is inherently challenging; such capability necessitates not only sophisticated model design but also considerable complexity. Interestingly, employing Gaussian noise as defense slightly increased the ASR for FFT attacks compared to SWAP, indicating that while noise affects high-frequency data, it doesn't completely remove relevant high-frequency information, merely masking it. If a model can denoise, it might still detect accurate high-frequency signals within the noise, implying that increasing perturbation frequency could enhance adversarial sample effectiveness, albeit with trade-offs to be considered in further analysis.

\vspace{-0.2cm}
\begin{figure*}[h]
    \centering
    \includegraphics[width=1.0\textwidth]{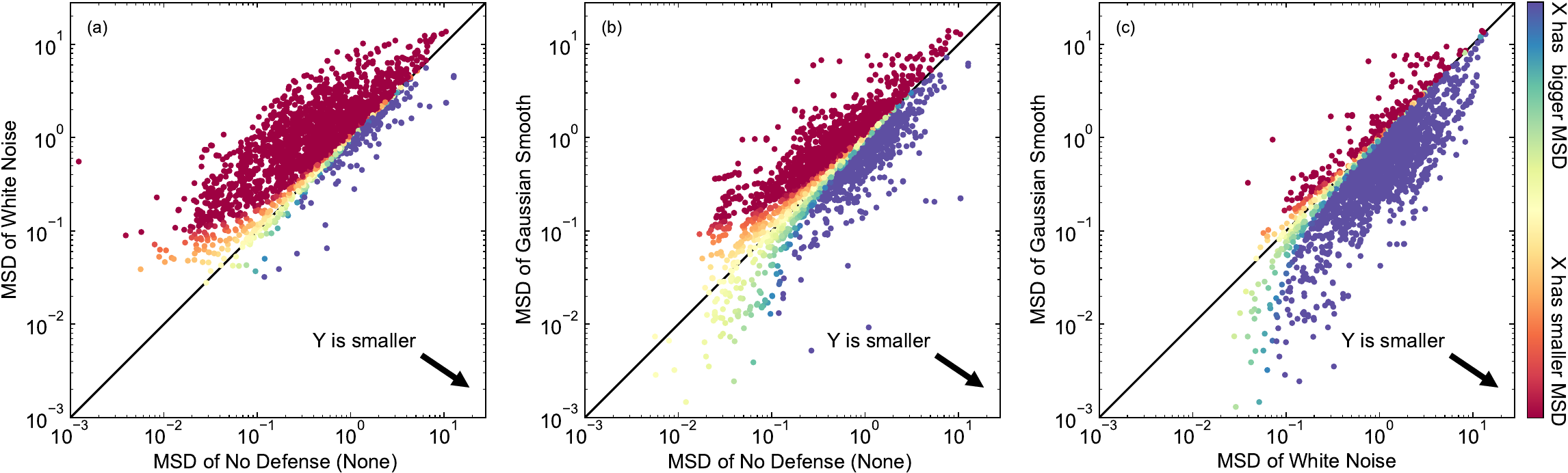}
    \vspace{-0.2cm}
    \caption{Comparison of Mean Success Distance (MSD) across different defense methods. Note that the data encompasses all attack methods and models presented herein.} \label{fig6}
\end{figure*}

Figure~\ref{fig6} presents a comparative analysis of the MSD across three data operation strategies, showcasing the MSD for five distinct models and five different attack methods under each augmentation strategy. Subgraph (a) distinctly illustrates an almost universal rise in MSD upon noise introduction, indicating that more significant perturbations are requisite for successful attacks. Contrarily, subgraph (b) reveals that the application of Gaussian filtering as a defense does not markedly alter MSD values in comparison to the unprotected scenarios. This nuanced observation is further elaborated in subgraph (c), delineating the disparate impacts of noise addition and low-pass filtering on MSD.

A meticulous examination of MSD values, especially under the threshold $r<0.1$, uncovers that models defended with Gaussian filtering exhibit significantly higher MSD compared to those either unprotected or shielded with noise. This trend, however, converges when larger perturbations \(r\) are necessary for a successful attack, highlighting a diminished discrepancy between Gaussian-blurred and undefended models. Such findings underscore that while both corrupting and eliminating high-frequency information escalate the complexity of launching an attack, the manifestation of these defensive measures diverges, offering insightful implications for designing robust defense mechanisms.

\vspace{-0.2cm}
\subsection{Analysis of attack and defense mechanisms}

\begin{figure}[h]
    \centering
    \includegraphics[width=1.0\textwidth]{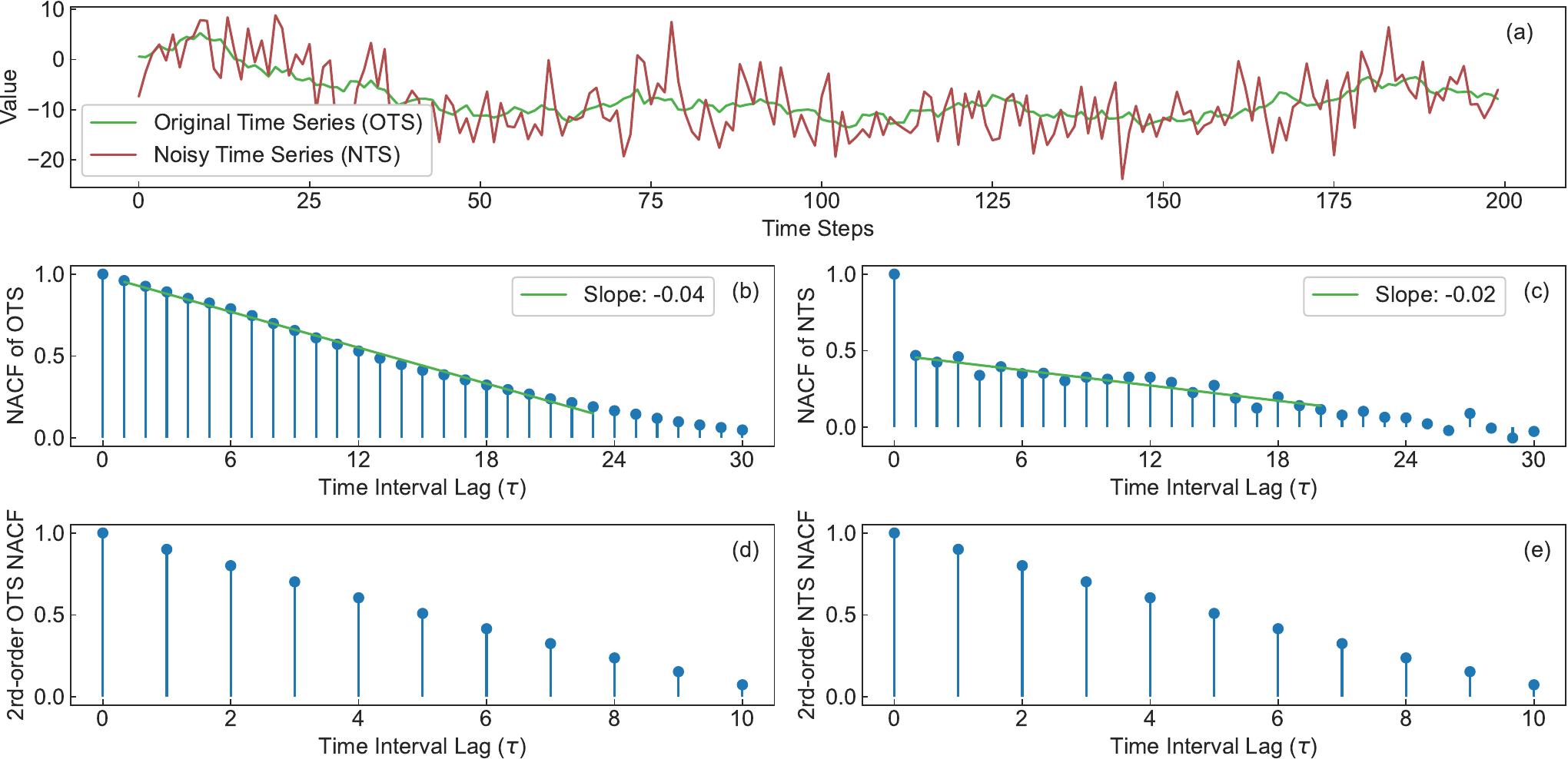}
    \caption{Illustration of the neural network how it adapts to learning global information through defense mechanisms, employing the Normalized Auto Correlation Function (NACF). It details key transformations: noise introduction transforming original time series~(OTS) into noisy ones~(NTS), and smoothing via high-frequency filtering, e.g., Gaussian filtering, reverting noisy series to their original state. These modifications significantly affect the NACF's correlations at small and large $\tau$ values, showcasing the resilience of long-distance correlations against local information distortion due to noise and filtering. Subfigures (b) and (c) depict the NACF's alignment after discarding the first data point, which the neural network automatically excludes as irrelevant to the label. Subfigures (d) and (e) demonstrate the consistency of the secondary autocorrelation function after the removal of contaminated points, underscoring the defense layer's effectiveness in maintaining pertinent global information amid disturbances.
    } \label{fig7}
\end{figure}

Figure~\ref{fig7} serves to elucidate the observed resilience of models against adversarial attacks upon the implementation of noise or filtering defenses. Subfigure (a) contrasts an original time series (OTS, in green) with its noisy counterpart (NTS, in red), while subfigures (b) and (c) analyze their auto correlations, respectively. The transformation from green to red symbolizes the effect of noise introduction, whereas Gaussian filtering essentially reverts the process. Crucially, this transformation highlights a significant disparity in correlation across small ($\tau=0$) and larger $\tau$ values within the NACF, suggesting a pronounced alteration in local correlations post-defense without substantially affecting larger scale correlations. Remarkably, the exclusion of the initial data point from the noise-affected time series-owing to its compromised local correlation-results in an autocorrelation function closely aligned along a straight line. Linear fitting post-removal further exemplifies the integrity of the correlations between the remaining data points. Moreover, the similarity in slopes between the two NACFs and the numerical ratio from the second to the first point underscores the minimal impact of defense on large-scale time series correlation, upon re-normalization.

In subfigures (d) and (e), we draw the autocorrelation function of the autocorrelation function. It can be seen that there is almost no difference between the two. This phenomenon, coupled with the intrinsic feature selection capability of deep neural networks, implies that the complexity of such networks suffices to discern and retain valuable global information despite local distortions. Consequently, this sheds light on why TS2V exhibits improved performance with defense augmentation and offers a broader understanding of defense mechanisms' efficacy in mitigating attack impacts. Interestingly, this analysis also suggests that simpler models may exhibit performance degradation (e.g., reduced test accuracy) upon defense layer addition, highlighting the differential learning challenges posed by local (high frequency) versus global (low frequency) information and providing insight into the effectiveness of high-frequency perturbations in adversarial attacks.

\vspace{-0.4cm}
\section{Conclusion}\label{sec:conclusion}
\vspace{-0.3cm}

In this paper, we explore the impact of adversarial attacks on time series classification models and assess the effectiveness of defense mechanisms, with a focus on the influence of frequency components on model vulnerability. Utilizing the Normalized Auto Correlation Function (NACF), we reveal a predilection within small neural networks for processing local rather than global information-a bias exploited by adversarial attacks. Our analysis shows that regularization strategies, especially those leveraging Fast Fourier Transform (FFT) methods, significantly improve attack success rates by emphasizing the frequency domain. In contrast, defenses like noise introduction and Gaussian filtering effectively diminish the Attack Success Rate (ASR), enhancing model robustness, particularly by mitigating high-frequency distortions. Notably, models designed to emphasize global information, such as the enhanced TS2V model, demonstrate superior resistance to adversarial manipulations. These results underscore the necessity of integrating considerations of local versus global information processing and frequency analysis in designing both attacks and defenses to bolster neural network resilience. This study advocates for further investigation into achieving an optimal balance between adversarial sophistication and model robustness, aiming to forge more secure and dependable machine learning frameworks.

\bibliographystyle{unsrt}

\end{document}